\documentclass[journal]{IEEEtran}
\usepackage{latexsym,,amssymb,amsmath,graphicx,epsf,cite,bbm,float}
\usepackage{ifpdf}
\usepackage{epstopdf,mathtools}

\usepackage{algorithm,algorithmic}
\usepackage{amsmath,amssymb,bm}
\usepackage{amsfonts,dsfont,color,bbm,subcaption}

\newcommand{\abs}[1]{|#1|}

\DeclareMathOperator*{\argmax}{arg\,max}
\DeclareMathOperator*{\argmin}{arg\,min}

\usepackage{hyperref,amsthm}

\newtheorem{theorem}{Theorem}

\newtheorem{lemma}[]{Lemma}

\newtheorem{proposition}[]{Proposition}

\newtheorem{corollary}[]{Corollary}

\newtheorem{remark}[]{Remark}

\newtheorem{definition}[]{Definition}

\begin{document}

\title{Smoothing with the Best Rectangle Window is Optimal for All Tapered Rectangle Windows} 
\author{\IEEEauthorblockN{Kaan Gokcesu}, \IEEEauthorblockN{Hakan Gokcesu} }
\maketitle

\begin{abstract}
	We investigate the optimal selection of weight windows for the problem of weighted least squares. We show that weight windows should be symmetric around its center, which is also its peak. We consider the class of tapered rectangle window weights, which are nonincreasing away from the center. We show that the best rectangle window is optimal for such window definitions. We also extend our results to the least absolutes and more general case of arbitrary loss functions to find similar results.
\end{abstract}

\section{Introduction}

In many machine learning problems, smoothing of a data set is desirable to detect behavioral patterns, and to filter out noise and outliers \cite{hardle1991smoothing}.
Fundamentally, smoothing works by the assumption that closer data points in some ordering sense (e.g., time index for temporal data) should be more closely related to each other. Henceforth, unexpectedly high valued observed data points should be lower and low ones should be higher.
Smoothing approaches, which are abundant in literature, can help in achieving a more flexible and robust analysis of datasets by constructing their more informative smoothed versions \cite{simonoff2012smoothing}. Because of the acquired more informative data, it has many applications in various fields from data analysis \cite{kenny1998data,brandt1998data,guo2004functional,deveaux1999applied,zin2020smoothing}, signal processing \cite{roberts1987digital,orfanidis1995introduction,fong2002monte,schaub2018flow}, anomaly detection \cite{chandola2009anomaly,gokcesu2017online,tsopelakos2019sequential, gokcesu2018sequential} and machine learning \cite{saul2004overview,gokcesu2020recursive,yaroshinsky2001smooth,gokcesu2020generalized,servedio2003smooth,gokcesu2021optimal,hartikainen2010kalman, gokcesu2021optimally}.

Smoothing differs with the closely-related field of curve-fitting \cite{arlinghaus1994practical}. While the goal in curve-fitting is to construct a functional form; smoothing is interested in just the smoothed values of dataset.
In smoothing, some parameter is generally used to determine the degree of smoothing; while in curve fitting, the inherent parameters are optimized for best fit.
The most traditional smoothers are linear smoothers \cite{buja1989linear}, where the smoothed data points are expressed in terms of other nearby data points with pre-determined linear weights. Hence, the smoothed signal is given by the convolution of the original signal with some finite impulse response (FIR) filter \cite{oppenheim1997signals}. 
Among linear smoothers, moving average is commonly used, especially with time-series data \cite{chatfield2013analysis,wei2006time,gokcesu2018adaptive,hansun2013new}. For the example of a temporal dataset, this approach attenuates short-term deviations and emphasizes the long-term behavior. The decision of what is considered short or long term, determines the moving average's input parameters (e.g., length) \cite{hatchett2010optimal}.
It is most prominently utilized in the field of mathematical finance, to analyze stock prices and trading volumes \cite{chiarella2006dynamic,gokcesu2021nonparametric}. It is also utilized in the analysis of many macroeconomic metrics such as GDP, import, export and unemployment \cite{schwert1987effects}. Moreover, it has applications in ECG analysis \cite{bai2004combination}, specifically in QRS detection \cite{chen2003moving}.
In mathematical terms, moving average is achieved by the convolution of the original signal by a probability distribution. Thus, it is a type of low-pass filter to filter out high frequency components, i.e., smooth out the signal \cite{kaiser1977data}.

There are other types of smoothing techniques aside from the moving average. One prominent example is the moving median \cite{justusson1981median}. Moving average is optimal in acquiring the true signal from a noisy observation, where the noise is Gaussian \cite{kalman1960new}. However, when the noise is non-Gaussian, specifically, the noise has a heavy tail, the performance degrades. For example, the moving average is susceptible to error in the presence of outliers. To this end, more robust approaches are desired \cite{gokcesu2021generalized,khargonekar1985non,gokcesu2022nonconvex}. One such example is moving median, which is a nonlinear smoother, where the smoothed values are acquired by the median of nearby samples, which can be efficiently computed by use of indexable skiplist \cite{pugh1990skip}. It is statistically optimal, when the noise on the true signal is distributed according to a Laplace distribution \cite{arce2005nonlinear}. This has application in many fields from edge detection \cite{huang1998computing} to mass spectrometry \cite{do1995applying}; especially in image processing, because of its edge-preserving smoothing abilities \cite{ataman1981some}. 

All in all, given an observation sequence $\{y_n\}_{n=1}^N$; smoothing techniques generally create the smoothed signal $\{x_n\}_{n=1}^N$ by using a sliding window \cite{lee2001sliding} of some length $2K+1$, i.e.,
\begin{align}
	x_n=\mathcal{T}(\{y_m\}_{m=n-K}^{n+K}),
\end{align}
where the function $\mathcal{T}(\cdot)$ can be mean, median or some other function altogether. A more generalized definition is achieved by the incorporation of a window function $\{w_k\}_{k=-K}^{K}$, where the smoothed signal is created as
\begin{align}
	x_n=\mathcal{T}(\{y_m,w_{m-n}\}_{m=n-K}^{n+K}),
\end{align}
where mean, median, etc. becomes weighted mean, weighted median, etc. Henceforth, the design of such window functions are just as important. 
%is a mathematical function that is zero-valued outside of some chosen interval, normally symmetric around the middle of the interval, usually near a maximum in the middle, and usually tapering away from the middle. 

Window functions \cite{weisstein2002crc} are utilized in spectral analysis \cite{stoica2005spectral} as well as antenna design \cite{rudge1982handbook} and beam-forming \cite{van1988beamforming}.
However, we mainly focus on their applications in the field of statistical analysis \cite{dixon1951introduction}, for the smoothing of a dataset, where they define a weighting vector. In curve fitting \cite{o1978curve} and in the field of Bayesian analysis \cite{ghosh2006introduction}, it is also referred as a kernel \cite{keerthi2003asymptotic}.
Kernel smoothing \cite{wand1994kernel} is a statistical technique to create data estimates as a weighted average (where weights are defined by a kernel) of its neighboring observations. This weighting is such that it diminishes from its peak in all directions, i.e., closer data points are given higher weights.
To this end, we tackle the problem of optimal kernel design, by considering the general class of tapered rectangle windows, where the peak is equally weighted with its immediate adjacent points and non-increasing further away.

%In \autoref{sec:problem}, we mathematically formalize problem of optimal weighted moving average smoothing. In \autoref{sec:quadratic}, we show that this optimization problem is a concave quadratic minimization problem. In \autoref{sec:polytope}, we provide an optimal solution when minimizing over a convex polytope. In \autoref{sec:window}, we prove that the best rectangle window is optimal for the class of tapered rectangle windows. In \autoref{sec:median}, we extend our results for the weighted moving median smoothing. In \autoref{sec:discussion}, we conclude with some discussions and remarks.

\section{Problem Description}\label{sec:problem}

Let us have the observed samples 
\begin{align}
	\boldsymbol{y}=\{y_n\}_{n=1}^{N}.
\end{align}
We smooth $\boldsymbol{y}$ and create our smoothed signal
\begin{align}
	\boldsymbol{x}=\{x_n\}_{n=1}^{N}.
\end{align}
We start by focusing on the design of a moving average smoother. Hence, to create $\boldsymbol{x}$, we pass $\boldsymbol{y}$ through a weighted moving average (weighted mean) filter $\boldsymbol{w}=\{w_k\}_{k=-M}^{M}$ of window size $2M+1\leq N$, where $\boldsymbol{w}$ is in a probability simplex. Hence, the weights are positive, i.e.,
\begin{align}
	w_k\geq 0,&&k\in\{-M,\ldots,M\},
\end{align}
and they sum to $1$, i.e.,
\begin{align}
	\sum_{k=-M}^{M}w_k=1.
\end{align}
Let $N$ be odd, i.e., $N=2K+1$ for some natural number $K$. Then, without loss of generality, we can define the weight vector $\boldsymbol{w}$ for a window length of $N=2K+1$ with some trailing zeros, i.e.,
\begin{align}
	\tilde{w}_k=\begin{cases}
		w_k,&-M\leq k\leq M\\
		0,& \text{otherwise}
	\end{cases},
\end{align} 
for $k\in\{-K,\ldots,K\}$. Thus, without loss of generality, we can assume the window length is $N$.
The smoothed signal $\boldsymbol{x}$ is given by
\begin{align}
	x_n=\sum_{k=-K}^{K}w_ky_{n+k},\label{eq:wmean}
\end{align}  
where the mean filter is cyclic, i.e., 
\begin{align}
	y_{n+k}=y_{(n+k-1)\pmod N+1}.\label{eq:ycyclic}
\end{align}
Note that the output $\boldsymbol{x}$ of this weighted moving average is a global minimizer of the following optimization problem (weighted cumulative cross error):
\begin{align}
	\min_{\boldsymbol{x}\in\Re^N}\sum_{n=1}^{N}\sum_{k=-K}^{K}w_k(y_{n+k}-x_n)^2,\label{eq:probx}
\end{align} 
since it is convex and the gradient is zero at its minimizer. Hence, given a weight vector $\boldsymbol{w}$, the solution of \eqref{eq:probx} is \eqref{eq:wmean}. 

However, a question arises about which $\boldsymbol{w}$ is most suitable for smoothing purposes given an observation vector $\boldsymbol{y}$. To find the optimal $\boldsymbol{w}$, we minimize the objective function in \eqref{eq:probx} with respect to $\boldsymbol{w}$, i.e.,
\begin{align}
	\argmin_{\boldsymbol{w}\in\mathcal{W}}\min_{\boldsymbol{x}\in\Re^N}\sum_{n=1}^{N}\sum_{k=-K}^{K}w_k(y_{n+k}-x_n)^2\label{eq:probw},
\end{align} 
where $\mathcal{W}$ is a subset of the probability simplex of dimension $N$, i.e., $\mathcal{P}^N$.
Later, we will specifically consider the set of tapered rectangle windows, where for every $\boldsymbol{{w}}\in\mathcal{W}$, we have
\begin{align}
	\boldsymbol{{w}}:
	\begin{cases}
		w_{i-1}=w_i=w_{i+1}&\\
		w_j\geq w_k,& i\leq j\leq k\\
		w_j\leq w_k,& j\leq k \leq i
	\end{cases},\label{eq:taper}
\end{align}
where the window $\boldsymbol{{w}}$ diminishes from its peak $w_i$.

\section{Formulation as a Quadratic Programming}\label{sec:quadratic}
Let the objective function $F(\boldsymbol{w},\boldsymbol{x})$ be defined as
\begin{align}
	F(\boldsymbol{w},\boldsymbol{x})=\sum_{n=1}^{N}\sum_{k=-K}^{K}w_k(y_{n+k}-x_n)^2,
\end{align}
where the optimization problem in \eqref{eq:probw} becomes
\begin{align}
	\argmin_{\boldsymbol{w}\in\mathcal{W}}\min_{\boldsymbol{x}\in\Re^N}F(\boldsymbol{w},\boldsymbol{x}).
\end{align}
Let us define the objective with respect to $\boldsymbol{w}$, $G(\boldsymbol{w})$ as
\begin{align}
	G(\boldsymbol{w})=\min_{\boldsymbol{x}\in\Re^N}F(\boldsymbol{w},\boldsymbol{x}),
\end{align}
where the optimization problem becomes
\begin{align}
	\argmin_{\boldsymbol{w}\in\mathcal{W}}G(\boldsymbol{w}).\label{eq:minG}
\end{align}
We can write $G(\boldsymbol{w})$ as follows:
\begin{align}
	G(\boldsymbol{w})=&\sum_{n=1}^{N}\sum_{k=-K}^{K}w_k\left(y_{n+k}-\sum_{m=-K}^{K}w_{m}y_{n+m}\right)^2.
\end{align}
\begin{definition}\label{def:r}
	The autocorrelation of $\boldsymbol{y}$ is defined as follows:
	\begin{align}
		r_{t}=\sum_{n=1}^{N}y_ny_{n+t},
	\end{align} 
	where ${y}_n$ is cyclic as in \eqref{eq:ycyclic}.
\end{definition}

\begin{proposition}\label{thm:rt}
	From \autoref{def:r} and \eqref{eq:ycyclic}, we have the following properties:
	\begin{itemize}
		\item $r_t$ is symmetric around $0$, i.e., $r_t=r_{-t}$.
		\item $r_t$ is periodic with $N$, i.e., $r_{t}=r_{N+t}$.
		\item For any integer $k$, we have $\sum_{n=1}^{N}y_{n+k}y_{n+k+t}=r_t$.
	\end{itemize}
\end{proposition}

\begin{lemma}\label{thm:quad}
	Using \autoref{thm:rt}, we can reduce the optimization problem to the following:
	\begin{align*}
		\argmax_{\boldsymbol{w}\in\mathcal{W}}\boldsymbol{w}^T\boldsymbol{R}\boldsymbol{w},
	\end{align*}
	where $\boldsymbol{w}$ is the weight vector and $\boldsymbol{R}$ is the autocorrelation matrix (i.e., $\boldsymbol{R}(i,j)=r_{j-i}$).
	\begin{proof}
		Using \autoref{thm:rt} and after some algebra, we can alternatively write the objective function $G(\boldsymbol{w})$ as
		\begin{align}
			G(\boldsymbol{w})%=&\sum_{k=-K}^{K}w_k\sum_{n=1}^{N}y_{n+k}^2\nonumber\\
			%&-2\sum_{k=-K}^{K}w_k\sum_{m=-K}^{K}w_m\sum_{n=1}^{N}y_{n+k}y_{n+m}\nonumber\\
			%&+\sum_{k=-K}^{K}w_{k}\sum_{m=-K}^{K}w_m\sum_{l=-K}^{K}w_{l}\sum_{n=1}^{N}y_{n+m}y_{n+l},\\
			=&\sum_{k=-K}^Kw_kr_0-2\sum_{k=-K}^{K}w_k\sum_{m=-K}^Kw_mr_{m-k}\nonumber\\
			&+\sum_{k=-K}^{K}w_k\sum_{m=-K}^Kw_m\sum_{k=-K}^{K}w_lr_{l-m},\\
			=&r_0-\sum_{k=-M}^{M}\sum_{m=-K}^{K}w_kw_mr_{m-k},\\
			=&r_0-\boldsymbol{w}^T\boldsymbol{R}\boldsymbol{w}.
		\end{align}
		We can change the objective in \eqref{eq:minG} to the following:
		\begin{align}
			\argmax_{\boldsymbol{w}\in\mathcal{W}} \left[r_0-G(\boldsymbol{w})\right],
		\end{align}
		which concludes the proof.
	\end{proof}
\end{lemma}

\section{Solution in a Convex Polytope}\label{sec:polytope}

We start by showing that the maximization problem in \autoref{thm:quad}, is a convex problem.
\begin{lemma}
	The autocorrelation matrix $R$ is positive semi-definite, i.e.,
	\begin{align*}
		\boldsymbol{v}^T\boldsymbol{R}\boldsymbol{v}\geq 0,
	\end{align*}
	for any $\boldsymbol{v}\in\Re^N$.
	\begin{proof}
		From \autoref{def:r}, we can write the matrix $\boldsymbol{{R}}$ as a sum of outer products
		\begin{align}
			\boldsymbol{R}=\sum_{n=1}^{N}\boldsymbol{z}_n\boldsymbol{z}_n^T,
		\end{align}
		where the vector $\boldsymbol{z}_n$ is equal to the vector $\boldsymbol{y}$ that is shifted by $n$ in a cyclic manner. Hence,
		\begin{align}
			\boldsymbol{v}^T\boldsymbol{R}\boldsymbol{v}=&\sum_{n=1}^{N}\boldsymbol{v}^T\boldsymbol{z}_n\boldsymbol{z}_n^T\boldsymbol{v},\\
			=&\sum_{n=1}^{N}\alpha_n^2,
		\end{align}
		where $\alpha_n$ is the inner product of $\boldsymbol{v}$ with $\boldsymbol{z}_n$. Since sum of squares are nonnegative, we conclude the proof.
	\end{proof}
\end{lemma}

Since $\boldsymbol{R}$ is positive semidefinite, the objective in \autoref{thm:quad} is the maximization of a convex function over the set $\mathcal{W}$. Let us consider the case that the set $\mathcal{W}$ is a convex polytope.

\begin{definition}\label{def:poly}
	Let the subset $\mathcal{W}$ be a convex polytope defined by its $I$ number of vertices $\boldsymbol{v_i}$ for $i\in\{1,\ldots,I\}$. Let the vertex matrix $\boldsymbol{V}$ be such that the $i^{th}$ column of $\boldsymbol{V}$ is $\boldsymbol{v_i}$.
\end{definition}

When $\mathcal{W}$ is a convex polytope as defined in \autoref{def:poly}, we have the following property.

\begin{proposition}\label{thm:poly}
	From \autoref{def:poly}, we can write any point $\boldsymbol{w}\in\mathcal{W}$ as a convex combination of the vertices $\{\boldsymbol{v_i}\}_{i=1}^I$, i.e.,
	\begin{align*}
		\boldsymbol{w}=\sum_{i=1}^{I}p_i\boldsymbol{v_i},
	\end{align*}
	where $\{p_i\}_{i=1}^I$ is in the probability simplex $\mathcal{P}^I$, i.e., 
	$p_i\geq 0$ for $ i\in\{1,\ldots,I\}$ and $\sum_{i=1}^{I}p_i=1$. Thus, we can write $\boldsymbol{w}$ as
	\begin{align*}
		\boldsymbol{w}=\boldsymbol{V}\boldsymbol{p},
	\end{align*}
	where the $i^{th}$ column of the matrix $\boldsymbol{V}$ is $\boldsymbol{v_i}$ and $\boldsymbol{p}=\{p_i\}_{i=1}^I$.
\end{proposition}

\begin{lemma}\label{thm:quadp}
	When the set $\mathcal{W}$ is a convex polytope with $I$ vertices defined by its vertex matrix $\boldsymbol{V}$ as in \autoref{def:poly}, we can reduce the optimization problem to the following:
	\begin{align*}
		\argmax_{\boldsymbol{p}\in\mathcal{P}^I}\boldsymbol{p}^T\boldsymbol{C}\boldsymbol{p},
	\end{align*}
	where $\boldsymbol{C}=\boldsymbol{V}^T\boldsymbol{R}\boldsymbol{V}$ and the set $\mathcal{P}^I$ is the probability simplex of dimension $I$.
	\begin{proof}
		The proof comes from rewriting the optimization problem in \autoref{thm:quad} with \autoref{thm:poly}.
	\end{proof}
\end{lemma}

For a convex maximization problem over a probability simplex, we have following fundamental result.
\begin{lemma}\label{thm:p1}
	A one-hot vector $\boldsymbol{p^*}=\{p^*_i\}_{i=1}^I$ (where $p^*_j=1$ and $p^*_i=0$ for $i\neq j$, for some $j\in\{1,\ldots,I\}$) is a maximizer of a convex function $H(\boldsymbol{p})$ over the probability simplex $\boldsymbol{p}\in\mathcal{P}^I$.
	\begin{proof}
		For any $\boldsymbol{p}\in\mathcal{P}^I$, we have
		\begin{align}
			\boldsymbol{p}=\sum_{i=1}^{I}p_i\boldsymbol{q_i},
		\end{align}
		where $\{p_i\}_{i=1}^I=\boldsymbol{p}$ and $\boldsymbol{q_i}$ is the one-hot vector whose $i^{th}$ element is $1$ and other elements are $0$. Thus, from Jensen inequality, we have
		\begin{align}
			H(\boldsymbol{p})\leq& \sum_{i=1}^{I}p_iH(\boldsymbol{q_i}),\\
			\leq& \max_{i\in\{1,\ldots,I\}}H(\boldsymbol{q_i}),
		\end{align}
		which concludes the proof.
	\end{proof}
\end{lemma}

Using \autoref{thm:p1} in conjunction with \autoref{thm:quadp}, we get the following theorem.
\begin{theorem} \label{thm:polyvertice}
	If the set $\mathcal{W}$ in the optimization problem of \autoref{thm:quad} is a convex polytope as in \autoref{def:poly}, then one of its vertices $\boldsymbol{v_i}$ is a maximizer, i.e.,
	\begin{align*}
		\boldsymbol{v_i}^T\boldsymbol{R}\boldsymbol{v_i}=\max_{\boldsymbol{w}\in\mathcal{W}}\boldsymbol{w}^T\boldsymbol{R}\boldsymbol{w},
	\end{align*}
	for some $i\in\{1,\ldots,I\}$.
	\begin{proof}
		When $\mathcal{W}$ is convex polytope as in \autoref{def:poly}, the optimization problem of \autoref{thm:quad} can be written as in \autoref{thm:quadp}. We see that $\boldsymbol{V}^T\boldsymbol{R}\boldsymbol{V}$ is positive semidefinite, i.e.,
		\begin{align}
			\boldsymbol{c}^T\boldsymbol{V}^T\boldsymbol{R}\boldsymbol{V}\boldsymbol{c}\geq 0,
		\end{align}
		for any $\boldsymbol{{c}}$, since $\boldsymbol{V}\boldsymbol{c}$, itself, is a vector and $\boldsymbol{R}$ is positive semidefinite. Hence, this is a convex maximization problem over a probability simplex. From \autoref{thm:p1}, we know that a one-hot vector for $\boldsymbol{p}$ is a maximizer. Since $\boldsymbol{w}=\boldsymbol{V}\boldsymbol{p}$, one of the vertices is a maximizer of the original problem, which concludes the proof.
	\end{proof}
\end{theorem}

We point out that a probability simplex itself is also a convex polytope. Hence, if the weights $\boldsymbol{w}$ can be any probability distribution of dimension $N$, we have the following result.
\begin{corollary}\label{thm:WPN}
	When $\mathcal{W}=\mathcal{P}^N$, a one-hot vector $\boldsymbol{w^*}$, i.e.,
	\begin{align*}
		w^*_k=\begin{cases}
			1,& k=i\\
			0,& k\neq0
		\end{cases},
	\end{align*}
	for some $i\in\{-K,\ldots,K\}$, where $\{w^*_k\}_{k=-K}^K=\boldsymbol{w^*}$ is an optimizer.
	\begin{proof}
		The proof comes from \autoref{thm:p1}.
	\end{proof}
\end{corollary}

Instead of the solution of \autoref{thm:WPN}, which provides no smoothing, we desire a more meaningful solution. Hence, we properly design the subset $\mathcal{W}$ in the next section.

\section{Designing the Window Function}\label{sec:window}
For the design of the window function, i.e., the set of weights $\mathcal{W}$, we start by considering some properties. 
\begin{lemma}
	Let the vector $\boldsymbol{w_n}$ be a shifted by $n$ (in a cyclic manner) version of the vector $\boldsymbol{w}$. Then, we have
	\begin{align*}
		\boldsymbol{w_n}^T\boldsymbol{R}\boldsymbol{w_n}=\boldsymbol{w}^T\boldsymbol{R}\boldsymbol{w},
	\end{align*}  
	for any integer $n$.
	\begin{proof}
		Cyclic shift of a vector $\boldsymbol{w}$ by $1$ can be achieved by multiplying by a matrix $\boldsymbol{S_1}$, whose rows are equal to the cyclic shift of the identity matrix by $1$. We observe that 
		\begin{align}
			\boldsymbol{S_1}^T\boldsymbol{R}\boldsymbol{S_1}=\boldsymbol{R},
		\end{align}
		since $\boldsymbol{R}$ is a circulant matrix. Let $\boldsymbol{S_n}$ be the matrix that achieves a cyclic shift of $n$, then we have
		\begin{align}
			\boldsymbol{S_n}=\boldsymbol{S_1}^n,
		\end{align} 
		where $\boldsymbol{S_1}^n$ is the successive multiplication of $\boldsymbol{S_1}$ by $n$ times. Hence,
		\begin{align}
			\boldsymbol{S_n}^T\boldsymbol{RS_n}=\boldsymbol{R},
		\end{align}
		and 
		\begin{align}
			\boldsymbol{w_n}^T\boldsymbol{R}\boldsymbol{w_n}=\boldsymbol{w}^T\boldsymbol{Rw},
		\end{align}
		which concludes the proof.
	\end{proof}
\end{lemma}

This lemma shows that any cyclic shift of the weight vector $\boldsymbol{w}$ will have the same objective value, which is intuitive since it would be equivalent to shifting the observation $\boldsymbol{y}$ itself. Thus, without loss generalization, we choose the solution which has its maximum at $w_0$, i.e., the weights have their peak at $k=0$. 

\begin{lemma}
	Let the vector $\boldsymbol{\tilde{w}}=\{\tilde{w}_k\}_{k=-K}^K$ be symmetric to the vector $\boldsymbol{w}=\{w_k\}_{k=-K}^K$ around $k=0$, i.e.,
	\begin{align*}
		\tilde{w}_k=w_{-k},
	\end{align*}
for $k\in\{-K,\ldots,K\}$. Then, we have
	\begin{align*}
		\boldsymbol{\tilde{w}}^T\boldsymbol{R\tilde{w}}=\boldsymbol{w}^T\boldsymbol{Rw}.
	\end{align*}
\begin{proof}
	We observe that
	\begin{align}
		\boldsymbol{\tilde{w}}=\boldsymbol{J}\boldsymbol{w},
	\end{align}
	where $\boldsymbol{J}$ is an exchange matrix (or row-reversed identity matrix), whose anti diagonal is all $1$ and $0$ otherwise. Hence,
	\begin{align}
		\boldsymbol{\tilde{w}}^T\boldsymbol{R\tilde{w}}=\boldsymbol{w}^T\boldsymbol{J}^T\boldsymbol{RJw}.
	\end{align}
	Since $\boldsymbol{R}$ is a symmetric circulant matrix, we have
	\begin{align}
		\boldsymbol{J}^T\boldsymbol{RJ}=\boldsymbol{R},
	\end{align}
	which gives
	\begin{align}
		\boldsymbol{\tilde{w}}^T\boldsymbol{R\tilde{w}}=\boldsymbol{w}^T\boldsymbol{Rw},
	\end{align}
	and concludes the proof.
\end{proof}
\end{lemma}

Since this result shows that two weights that are symmetric around $k=0$ to each other have the same loss, we consider the set of weights that are symmetric around $k=0$.

For the set of symmetric window distributions, we define the following.
\begin{definition}\label{def:Wsymm}
	Let us define $\mathcal{W}_{symm}$ as the set weights $\boldsymbol{w}=\{w_k\}_{k=-K}^K$ that satisfy
	\begin{align*}
		w_0=&(1+\epsilon)w_1,\\
		w_{k}=&w_{-k},&&\forall k\\
		w_{k}\geq&w_{k+1},&&k\geq1,\\
		w_{k}\geq&w_{k-1},&&k\leq-1,
	\end{align*}
	for some $\epsilon\geq0$.
\end{definition}
Note that we have constrained $w_0$, because otherwise, the optimal solution would have been a one-hot vector. When $\epsilon=0$, we have the set of tapered rectangle windows in \eqref{eq:taper}.
\begin{lemma}\label{Wsymmvertice}
	The set $\mathcal{W}_{symm}$ defined in \autoref{def:Wsymm} is a convex polytope with $K$ vertices $\{\boldsymbol{v_i}\}_{i=1}^K$, where $\boldsymbol{v_i}=\{v_{i,k}\}_{k=-K}^K$ is given by
	\begin{align*}
		v_{i,k}=\begin{cases}
			\frac{1+\epsilon}{2i+1+\epsilon},& k=0\\
			\frac{1}{2i+1+\epsilon},& 1\leq\abs{k}\leq i\\
			0,& \text{otherwise}
		\end{cases},
	\end{align*}
	for some parameter $\epsilon\geq0$.
	\begin{proof}
		Any $\boldsymbol{w}\in \mathcal{W}_{symm}$ can be written as
		\begin{align}
			\boldsymbol{w}=\sum_{i=1}^{K}p_i\boldsymbol{v_i},
		\end{align}
		where 
		\begin{align}
			p_i=\begin{cases}
				w_K(2K+1+\epsilon),& i=K\\
				(w_i-w_{i+1})(2i+1+\epsilon),& {1\leq i\leq K-1}
			\end{cases}.
		\end{align}
		Let us define $w_{K+1}\triangleq0$, we have
		\begin{align}
			\sum_{i=1}^Kp_i=&\sum_{i=1}^{K}(w_i-w_{i+1})(2i+1+\epsilon),\\
			=&\sum_{i=1}^{K}(1+\epsilon)(w_i-w_{i+1})+2\sum_{i=1}^{K}\sum_{j=i}^{K}(w_j-w_{j+1}),\nonumber\\
			=&(1+\epsilon)w_1+2\sum_{i=1}^{K}w_i,\\
			=&\sum_{i=-K}^{K}w_k
			=1,
		\end{align}
		since $w_0=(1+\epsilon)w_1$ and $w_i=w_{-i}$ from \autoref{def:Wsymm}. Moreover, $p_i\geq0$, which gives $\boldsymbol{p}=\{p_i\}_{i=-K}^K$ is a probability vector. From \autoref{thm:poly}, $\mathcal{W}_{symm}$ is a convex polytope defined by its vertices $\{\boldsymbol{v_i}\}_{i=1}^K$ for any $\epsilon\geq0$, which concludes the proof.
	\end{proof}
\end{lemma}

\begin{corollary}\label{thm:meanfilt}
	When $\epsilon=0$, an unweighted moving average of some odd length $i\geq 3$ is an optimizer for the set $\mathcal{W}_{symm}$ in \autoref{def:Wsymm}.
	\begin{proof}
		From \autoref{Wsymmvertice}, we see that, when $\epsilon=0$, $\mathcal{W}_{symm}$ is a convex polytope whose vertices consist of the moving average vectors with odd lengths of at least $3$. From \autoref{thm:polyvertice}, one of these vertices is an optimizer, which concludes the proof.
	\end{proof}
\end{corollary}

\section{Extension to Median Filter}\label{sec:median}
We observe that our derivations are not limited to the weighted mean filter and can also be applied to the weighted median filter as well.
For a given observation vector $\boldsymbol{y}=\{y_n\}_{n=1}^N$ and probability weights $\boldsymbol{w}=\{w_k\}_{k=-K}^K$, let the output $\boldsymbol{x}=\{x_n\}_{n=1}^N$ be created as follows:
\begin{align}
	x_n=\text{wmedian}\left(\{y_{n+k},w_k\}_{k=-K}^K\right),\label{eq:xwmedian}
\end{align}
where the function $\text{wmedian}(\cdot)$ orders $\{y_{n+k},w_{k}\}_{k=-K}^K$ according to $\{y_{n+k}\}_{k=-K}^K$. Let $\{\tilde{y}_{n+k}\}_{k=-K}^K$ be the ordered version of $\{y_{n+k}\}_{k=-K}^K$ and $\{\tilde{w}_k\}_{k=-K}^K$ be its corresponding re-indexed weights. Then, we have 
\begin{align}
	x_n=\tilde{y}_i, &&\exists i:
	\sum_{k=-K}^{i-1}\tilde{w}_k\leq 0.5,\sum_{k=i+1}^{K}\tilde{w}_k\leq 0.5.
\end{align}

\begin{definition}\label{def:F_A}
	$\boldsymbol{x}=\{x_n\}_{n=1}^N$ in \eqref{eq:xwmedian} is a solution to the optimization problem
	\begin{align*}
		\min_{\boldsymbol{x}\in\Re^N}F_A(\boldsymbol{w},\boldsymbol{x})\triangleq\min_{\boldsymbol{x}\in\Re^N}\sum_{n=1}^{N}\sum_{k=-K}^{K}w_k\abs{y_{n+k}-x_n}.
	\end{align*} 
\end{definition}

\begin{remark}
	The objective function $F_A(\boldsymbol{w},\boldsymbol{x})$ in \autoref{def:F_A}  is linear in its argument $\boldsymbol{w}$ for any fixed $\boldsymbol{x}\in\Re^N$.
\end{remark}

\begin{proposition}\label{thm:GAconc}
	The function $G_A(\boldsymbol{w})$, which is the minimization of $F_A(\boldsymbol{w},\boldsymbol{x})$ (in \autoref{def:F_A}) over $\boldsymbol{x}$, i.e., 
	\begin{align*}
		G_A(\boldsymbol{w})\triangleq\min_{\boldsymbol{x}\in\Re^N}F_A(\boldsymbol{w},\boldsymbol{x})
	\end{align*}
	is concave in $\boldsymbol{w}$.
	\begin{proof}
		Since $F_A(\boldsymbol{w},\boldsymbol{x})$ is linear, we have
		\begin{align}
			G_A(\lambda&\boldsymbol{w_0}+(1-\lambda)\boldsymbol{w_1})\\
			%=&\min_{\boldsymbol{x}\in\Re^N}F_A(\lambda\boldsymbol{w_0}+(1-\lambda)\boldsymbol{w_1},\boldsymbol{x}),\\
			=&\min_{\boldsymbol{x}\in\Re^N}\left[\lambda F_A(\boldsymbol{w_0},\boldsymbol{x})+(1-\lambda) F_A(\boldsymbol{w_1},\boldsymbol{x})\right],\\
			\geq&\min_{\boldsymbol{x}\in\Re^N}\lambda F_A(\boldsymbol{w_0},\boldsymbol{x})+\min_{\boldsymbol{x}\in\Re^N}(1-\lambda) F_A(\boldsymbol{w_1},\boldsymbol{x}),\\	
			\geq&\lambda G_A(\boldsymbol{w_0})+(1-\lambda)G_A(\boldsymbol{w_1}),
		\end{align}
		for any $0\leq\lambda\leq 1$, which concludes the proof.
	\end{proof}
\end{proposition}

\begin{remark}\label{rem:objA}
	We have the following concave minimization:
	\begin{align*}
		\min_{\boldsymbol{w}\in\mathcal{W}}G_A(\boldsymbol{w}).
	\end{align*}
\end{remark}

\begin{theorem} \label{thm:polyverticeA}
	If the set $\mathcal{W}$ in the optimization problem of \autoref{rem:objA} is a convex polytope defined by its set of vertices $\{\boldsymbol{v_i}\}_{i=1}^I$ as in \autoref{def:poly}, then one of its vertices $\boldsymbol{v}_j$ is a minimizer, i.e.,
	\begin{align*}
		G_A(\boldsymbol{v_j})=\min_{\boldsymbol{w}\in\mathcal{W}}G_A(\boldsymbol{w}),
	\end{align*}
	for some $j\in\{1,\ldots,I\}$.
	\begin{proof}
		The proof is similar as in \autoref{thm:polyvertice} and comes from \autoref{thm:p1}, 
		since $G_A(\boldsymbol{w})$ is concave from \autoref{thm:GAconc}.
	\end{proof}
\end{theorem}

\begin{corollary}\label{thm:medianfilt}
	For the set $\mathcal{W}_{symm}$ of \autoref{def:Wsymm}, when $\epsilon=0$, an unweighted median filter of some odd length $i\geq 3$ is an optimizer for \autoref{rem:objA}.
	\begin{proof}
		The proof is similar to the proof of \autoref{thm:meanfilt}.
	\end{proof}
\end{corollary}

%Hence, similar to the weighted mean filters, the optimal weighted median filter in $\mathcal{W}_{symm}$ is a simple median filter. 

\section{Discussions and Conclusion}\label{sec:discussion}

Similar to the case with the absolute loss, we observe that any arbitrary loss function has the same result. Let $\boldsymbol{x}$ be such that it is a minimization of the following objective function
\begin{align}
	\min_{\boldsymbol{x}\in\Re^N}F_G(\boldsymbol{w},\boldsymbol{x}),
\end{align}
where
\begin{align}
	F_G(\boldsymbol{w},\boldsymbol{x})\triangleq\sum_{n=1}^{N}\sum_{k=-K}^{K}w_kf(y_{n+k},x_n),
\end{align} 
for some function $f(\cdot,\cdot)$.

We observe that irrespective of the function $f(\cdot,\cdot)$, $F_G(\boldsymbol{w},\boldsymbol{x})$ is linear in $\boldsymbol{w}$. Thus,
We have 
\begin{align}
	G_G(\boldsymbol{w})=\min_{\boldsymbol{x}\in\Re^N}F_G(\boldsymbol{w},\boldsymbol{x}),
\end{align}
which is also concave in $\boldsymbol{w}$. Similarly to the median filter setting, when $\mathcal{W}$ is a convex polytope, one of its vertices is an optimizer.

In general, to find an optimizer in the convex polytope $\mathcal{W}_{symm}$, all of the vertices need to be tried to find the one with the smaller loss $G_G(\boldsymbol{w})$. For the median filtering problem, this has a polynomial in $N$ complexity. 
However, for the weighted mean filtering problem, it is much more efficient.

\begin{remark}
	For the quadratic optimization problem in \autoref{thm:polyvertice}, the optimal weights in $\mathcal{W}_{symm}$ can be found in linearithmic, i.e., $O(N\log N)$, complexity.
	\begin{proof}
		 From Wiener–Khinchin theorem \cite{chatfield2013analysis}, we can compute a row of the autocorrelation matrix $\boldsymbol{R}$ in $O(N\log N)$ time with two Fast-Fourier Transforms (FFT) \cite{heideman1985gauss}, since
		 \begin{align}
		 	\boldsymbol{r}=IFFT(\boldsymbol{f_y}\odot \boldsymbol{f_y^*}),
		 \end{align}
	 	where $\boldsymbol{r}$ is the autocorrelation vector (a single row of $\boldsymbol{R}$), the operation $\odot$ is element-wise multiplication, the super-script star denotes the conjugate and
	 	\begin{align}
	 		\boldsymbol{f_y}=FFT(\boldsymbol{y}).
	 	\end{align}
 		After finding the autocorrelation, starting with the length $3$ weight vector, the evaluation of the objective function in \autoref{thm:quad} can be recursively computed in $O(1)$ time for every odd length weight vector. Hence, the evaluation process has linear time complexity, i.e., $O(N)$ complexity and the total complexity is linearithmic, i.e., $O(N\log N)$. 
	\end{proof}
\end{remark}

In conclusion, we have investigated the optimal selection of weight windows for the problem of weighted least squares. We have shown that weight windows should be symmetric around its center, which is also the peak of the window. Moreover, we have considered that the weights should be discounted with sample distances; and assumed that the weights are nonincreasing away from the center (as per the tapered rectangle window definition). Surprisingly, for such window definitions, an optimal weighting has been shown to be an unweighted window of some odd size at least $3$. We have also extended our results to the least absolutes and the more general case of arbitrary loss functions, where we found similar results.

%\newpage
\bibliographystyle{IEEEtran}
\bibliography{double_bib}

\begin{thebibliography}{10}
\providecommand{\url}[1]{#1}
\csname url@samestyle\endcsname
\providecommand{\newblock}{\relax}
\providecommand{\bibinfo}[2]{#2}
\providecommand{\BIBentrySTDinterwordspacing}{\spaceskip=0pt\relax}
\providecommand{\BIBentryALTinterwordstretchfactor}{4}
\providecommand{\BIBentryALTinterwordspacing}{\spaceskip=\fontdimen2\font plus
\BIBentryALTinterwordstretchfactor\fontdimen3\font minus
  \fontdimen4\font\relax}
\providecommand{\BIBforeignlanguage}[2]{{%
\expandafter\ifx\csname l@#1\endcsname\relax
\typeout{** WARNING: IEEEtran.bst: No hyphenation pattern has been}%
\typeout{** loaded for the language `#1'. Using the pattern for}%
\typeout{** the default language instead.}%
\else
\language=\csname l@#1\endcsname
\fi
#2}}
\providecommand{\BIBdecl}{\relax}
\BIBdecl

\bibitem{hardle1991smoothing}
W.~K. H{\"a}rdle \emph{et~al.}, \emph{Smoothing techniques: with implementation
  in S}.\hskip 1em plus 0.5em minus 0.4em\relax Springer Science \& Business
  Media, 1991.

\bibitem{simonoff2012smoothing}
J.~S. Simonoff, \emph{Smoothing methods in statistics}.\hskip 1em plus 0.5em
  minus 0.4em\relax Springer Science \& Business Media, 2012.

\bibitem{kenny1998data}
D.~Kenny, D.~Kashy, and N.~Bolger, ``Data analysis,'' in \emph{The handbook of
  social psychology: Vols. 1 and 2,}.\hskip 1em plus 0.5em minus 0.4em\relax
  McGraw-Hill New York, 1998, pp. 233--265.

\bibitem{brandt1998data}
S.~Brandt and S.~Brandt, \emph{Data analysis}.\hskip 1em plus 0.5em minus
  0.4em\relax Springer, 1998.

\bibitem{guo2004functional}
W.~Guo, ``Functional data analysis in longitudinal settings using smoothing
  splines,'' \emph{Statistical methods in medical research}, vol.~13, no.~1,
  pp. 49--62, 2004.

\bibitem{deveaux1999applied}
R.~D. Deveaux, ``Applied smoothing techniques for data analysis,'' 1999.

\bibitem{zin2020smoothing}
M.~A.~M. Zin, A.~S. Rambely, N.~M. Ariff, and M.~S. Ariffin, ``Smoothing and
  differentiation of kinematic data using functional data analysis approach: an
  application of automatic and subjective methods,'' \emph{Applied Sciences},
  vol.~10, no.~7, p. 2493, 2020.

\bibitem{roberts1987digital}
R.~A. Roberts and C.~T. Mullis, \emph{Digital signal processing}.\hskip 1em
  plus 0.5em minus 0.4em\relax Addison-Wesley Longman Publishing Co., Inc.,
  1987.

\bibitem{orfanidis1995introduction}
S.~J. Orfanidis, \emph{Introduction to signal processing}.\hskip 1em plus 0.5em
  minus 0.4em\relax Prentice-Hall, Inc., 1995.

\bibitem{fong2002monte}
W.~Fong, S.~J. Godsill, A.~Doucet, and M.~West, ``Monte carlo smoothing with
  application to audio signal enhancement,'' \emph{IEEE transactions on signal
  processing}, vol.~50, no.~2, pp. 438--449, 2002.

\bibitem{schaub2018flow}
M.~T. Schaub and S.~Segarra, ``Flow smoothing and denoising: Graph signal
  processing in the edge-space,'' in \emph{2018 IEEE Global Conference on
  Signal and Information Processing (GlobalSIP)}.\hskip 1em plus 0.5em minus
  0.4em\relax IEEE, 2018, pp. 735--739.

\bibitem{chandola2009anomaly}
V.~Chandola, A.~Banerjee, and V.~Kumar, ``Anomaly detection: A survey,''
  \emph{ACM computing surveys (CSUR)}, vol.~41, no.~3, pp. 1--58, 2009.

\bibitem{gokcesu2017online}
K.~Gokcesu and S.~S. Kozat, ``Online anomaly detection with minimax optimal
  density estimation in nonstationary environments,'' \emph{IEEE Transactions
  on Signal Processing}, vol.~66, no.~5, pp. 1213--1227, 2017.

\bibitem{tsopelakos2019sequential}
A.~Tsopelakos, G.~Fellouris, and V.~V. Veeravalli, ``Sequential anomaly
  detection with observation control,'' in \emph{2019 IEEE International
  Symposium on Information Theory (ISIT)}.\hskip 1em plus 0.5em minus
  0.4em\relax IEEE, 2019, pp. 2389--2393.

\bibitem{gokcesu2018sequential}
K.~Gokcesu, M.~M. Neyshabouri, H.~Gokcesu, and S.~S. Kozat, ``Sequential
  outlier detection based on incremental decision trees,'' \emph{IEEE
  Transactions on Signal Processing}, vol.~67, no.~4, pp. 993--1005, 2018.

\bibitem{saul2004overview}
S.~Carliner, ``An overview of online learning (2nd ed.),'' \emph{European
  Business Review}, vol.~16, 01 2004.

\bibitem{gokcesu2020recursive}
K.~Gokcesu and H.~Gokcesu, ``Recursive experts: An efficient optimal mixture of
  learning systems in dynamic environments,'' \emph{arXiv preprint
  arXiv:2009.09249}, 2020.

\bibitem{yaroshinsky2001smooth}
R.~Yaroshinsky and R.~El-Yaniv, \emph{Smooth online learning of expert
  advice}.\hskip 1em plus 0.5em minus 0.4em\relax Citeseer, 2001.

\bibitem{gokcesu2020generalized}
K.~Gokcesu and H.~Gokcesu, ``A generalized online algorithm for translation and
  scale invariant prediction with expert advice,'' \emph{arXiv preprint
  arXiv:2009.04372}, 2020.

\bibitem{servedio2003smooth}
R.~A. Servedio, ``Smooth boosting and learning with malicious noise,''
  \emph{The Journal of Machine Learning Research}, vol.~4, pp. 633--648, 2003.

\bibitem{gokcesu2021optimal}
K.~Gokcesu and H.~Gokcesu, ``Optimal and efficient algorithms for general
  mixable losses against switching oracles,'' \emph{arXiv preprint
  arXiv:2108.06411}, 2021.

\bibitem{hartikainen2010kalman}
J.~Hartikainen and S.~S{\"a}rkk{\"a}, ``Kalman filtering and smoothing
  solutions to temporal gaussian process regression models,'' in \emph{2010
  IEEE international workshop on machine learning for signal processing}.\hskip
  1em plus 0.5em minus 0.4em\relax IEEE, 2010, pp. 379--384.

\bibitem{gokcesu2021optimally}
K.~Gokcesu and H.~Gokcesu, ``Optimally efficient sequential calibration of
  binary classifiers to minimize classification error,'' \emph{arXiv preprint
  arXiv:2108.08780}, 2021.

\bibitem{arlinghaus1994practical}
S.~Arlinghaus, \emph{Practical handbook of curve fitting}.\hskip 1em plus 0.5em
  minus 0.4em\relax CRC press, 1994.

\bibitem{buja1989linear}
A.~Buja, T.~Hastie, and R.~Tibshirani, ``Linear smoothers and additive
  models,'' \emph{The Annals of Statistics}, pp. 453--510, 1989.

\bibitem{oppenheim1997signals}
A.~V. Oppenheim, A.~S. Willsky, S.~H. Nawab, G.~M. Hern{\'a}ndez \emph{et~al.},
  \emph{Signals \& systems}.\hskip 1em plus 0.5em minus 0.4em\relax Pearson
  Educaci{\'o}n, 1997.

\bibitem{chatfield2013analysis}
C.~Chatfield, \emph{The analysis of time series: theory and practice}.\hskip
  1em plus 0.5em minus 0.4em\relax Springer, 2013.

\bibitem{wei2006time}
W.~W. Wei, ``Time series analysis,'' in \emph{The Oxford Handbook of
  Quantitative Methods in Psychology: Vol. 2}, 2006.

\bibitem{gokcesu2018adaptive}
K.~Gokcesu, M.~Ergeneci, E.~Ertan, and H.~Gokcesu, ``An adaptive algorithm for
  online interference cancellation in emg sensors,'' \emph{IEEE Sensors
  Journal}, vol.~19, no.~1, pp. 214--223, 2018.

\bibitem{hansun2013new}
S.~Hansun, ``A new approach of moving average method in time series analysis,''
  in \emph{2013 conference on new media studies (CoNMedia)}.\hskip 1em plus
  0.5em minus 0.4em\relax IEEE, 2013, pp. 1--4.

\bibitem{hatchett2010optimal}
R.~B. Hatchett, B.~W. Brorsen, and K.~B. Anderson, ``Optimal length of moving
  average to forecast futures basis,'' \emph{Journal of Agricultural and
  Resource Economics}, pp. 18--33, 2010.

\bibitem{chiarella2006dynamic}
C.~Chiarella, X.-Z. He, and C.~Hommes, ``A dynamic analysis of moving average
  rules,'' \emph{Journal of Economic Dynamics and Control}, vol.~30, no. 9-10,
  pp. 1729--1753, 2006.

\bibitem{gokcesu2021nonparametric}
K.~Gokcesu and H.~Gokcesu, ``Nonparametric extrema analysis in time series for
  envelope extraction, peak detection and clustering,'' \emph{arXiv preprint
  arXiv:2109.02082}, 2021.

\bibitem{schwert1987effects}
G.~W. Schwert, ``Effects of model specification on tests for unit roots in
  macroeconomic data,'' \emph{Journal of monetary economics}, vol.~20, no.~1,
  pp. 73--103, 1987.

\bibitem{bai2004combination}
Y.-W. Bai, W.-Y. Chu, C.-Y. Chen, Y.-T. Lee, Y.-C. Tsai, and C.-H. Tsai, ``The
  combination of kaiser window and moving average for the low-pass filtering of
  the remote ecg signals,'' in \emph{Proceedings. 17th IEEE Symposium on
  Computer-Based Medical Systems}.\hskip 1em plus 0.5em minus 0.4em\relax IEEE,
  2004, pp. 273--278.

\bibitem{chen2003moving}
H.~Chen and S.-W. Chen, ``A moving average based filtering system with its
  application to real-time qrs detection,'' in \emph{Computers in Cardiology,
  2003}.\hskip 1em plus 0.5em minus 0.4em\relax IEEE, 2003, pp. 585--588.

\bibitem{kaiser1977data}
J.~Kaiser and W.~Reed, ``Data smoothing using low-pass digital filters,''
  \emph{Review of Scientific Instruments}, vol.~48, no.~11, pp. 1447--1457,
  1977.

\bibitem{justusson1981median}
B.~Justusson, ``Median filtering: Statistical properties,'' in
  \emph{Two-Dimensional Digital Signal Prcessing II}.\hskip 1em plus 0.5em
  minus 0.4em\relax Springer, 1981, pp. 161--196.

\bibitem{kalman1960new}
R.~E. Kalman and Others, ``A new approach to linear filtering and prediction
  problems,'' \emph{Journal of basic Engineering}, vol.~82, no.~1, pp. 35--45,
  1960.

\bibitem{gokcesu2021generalized}
K.~Gokcesu and H.~Gokcesu, ``Generalized huber loss for robust learning and its
  efficient minimization for a robust statistics,'' \emph{arXiv preprint
  arXiv:2108.12627}, 2021.

\bibitem{khargonekar1985non}
P.~Khargonekar and A.~Tannenbaum, ``Non-euclidian metrics and the robust
  stabilization of systems with parameter uncertainty,'' \emph{IEEE
  Transactions on Automatic Control}, vol.~30, no.~10, pp. 1005--1013, 1985.

\bibitem{gokcesu2022nonconvex}
K.~Gokcesu and H.~Gokcesu, ``Nonconvex extension of generalized huber loss for
  robust learning and pseudo-mode statistics,'' \emph{arXiv preprint
  arXiv:2202.11141}, 2022.

\bibitem{pugh1990skip}
W.~Pugh, ``Skip lists: a probabilistic alternative to balanced trees,''
  \emph{Communications of the ACM}, vol.~33, no.~6, pp. 668--676, 1990.

\bibitem{arce2005nonlinear}
G.~R. Arce, \emph{Nonlinear signal processing: a statistical approach}.\hskip
  1em plus 0.5em minus 0.4em\relax John Wiley \& Sons, 2005.

\bibitem{huang1998computing}
D.~Huang and W.~T. Dunsmuir, ``Computing joint distributions of 2d moving
  median filters with applications to detection of edges,'' \emph{IEEE
  transactions on pattern analysis and machine intelligence}, vol.~20, no.~3,
  pp. 340--343, 1998.

\bibitem{do1995applying}
C.~L. do~Lago, V.~F. Juliano, and C.~Kascheres, ``Applying moving median
  digital filter to mass spectrometry and potentiometric titration,''
  \emph{Analytica Chimica Acta}, vol. 310, no.~2, pp. 281--288, 1995.

\bibitem{ataman1981some}
E.~Ataman, V.~Aatre, and K.~Wong, ``Some statistical properties of median
  filters,'' \emph{IEEE Transactions on Acoustics, Speech, and Signal
  Processing}, vol.~29, no.~5, pp. 1073--1075, 1981.

\bibitem{lee2001sliding}
C.-H. Lee, C.-R. Lin, and M.-S. Chen, ``Sliding-window filtering: an efficient
  algorithm for incremental mining,'' in \emph{Proceedings of the tenth
  international conference on Information and knowledge management}, 2001, pp.
  263--270.

\bibitem{weisstein2002crc}
E.~W. Weisstein, \emph{CRC concise encyclopedia of mathematics}.\hskip 1em plus
  0.5em minus 0.4em\relax Chapman and Hall/CRC, 2002.

\bibitem{stoica2005spectral}
P.~Stoica and R.~Moses, \emph{Spectral Analysis of Signals}.\hskip 1em plus
  0.5em minus 0.4em\relax Pearson Prentice Hall, 2005.

\bibitem{rudge1982handbook}
A.~W. Rudge and K.~Milne, \emph{The handbook of antenna design}.\hskip 1em plus
  0.5em minus 0.4em\relax Iet, 1982, vol.~16.

\bibitem{van1988beamforming}
B.~D. Van~Veen and K.~M. Buckley, ``Beamforming: A versatile approach to
  spatial filtering,'' \emph{IEEE assp magazine}, vol.~5, no.~2, pp. 4--24,
  1988.

\bibitem{dixon1951introduction}
W.~J. Dixon and F.~J. Massey~Jr, \emph{Introduction to statistical
  analysis.}\hskip 1em plus 0.5em minus 0.4em\relax McGraw-Hill, 1951.

\bibitem{o1978curve}
A.~O'Hagan, ``Curve fitting and optimal design for prediction,'' \emph{Journal
  of the Royal Statistical Society: Series B (Methodological)}, vol.~40, no.~1,
  pp. 1--24, 1978.

\bibitem{ghosh2006introduction}
J.~K. Ghosh, M.~Delampady, and T.~Samanta, \emph{An introduction to Bayesian
  analysis: theory and methods}.\hskip 1em plus 0.5em minus 0.4em\relax
  Springer, 2006, vol. 725.

\bibitem{keerthi2003asymptotic}
S.~S. Keerthi and C.-J. Lin, ``Asymptotic behaviors of support vector machines
  with gaussian kernel,'' \emph{Neural computation}, vol.~15, no.~7, pp.
  1667--1689, 2003.

\bibitem{wand1994kernel}
M.~P. Wand and M.~C. Jones, \emph{Kernel smoothing}.\hskip 1em plus 0.5em minus
  0.4em\relax CRC press, 1994.

\bibitem{heideman1985gauss}
M.~T. Heideman, D.~H. Johnson, and C.~S. Burrus, ``Gauss and the history of the
  fast fourier transform,'' \emph{Archive for history of exact sciences}, pp.
  265--277, 1985.

\end{thebibliography}

\end{document}